\documentclass[sigconf]{acmart}

\pdfoutput=1

\usepackage{multirow}
\usepackage[american]{babel}
\usepackage{amsmath,amsfonts}
\usepackage{microtype}
\usepackage{graphicx}
\usepackage{booktabs}
\usepackage{subfigure}
\graphicspath{{figure/}}
\begin{document}

\title{
    Beyond Statistical Relations: Integrating Knowledge Relations into Style Correlations for Multi-Label Music Style Classification
}



\author{
	Qianwen Ma, 
	Chunyuan Yuan
}
\email{{maqianwen,yuanchunyuan}@iie.ac.cn}
\affiliation{
	\institution{Institute of Information Engineering, Chinese Academy of Sciences}
	\institution{School of Cyber Security, University of Chinese Academy of Sciences}
}
\authornote{Both authors contributed equally to this research.}

\author{
	Wei Zhou, 
	Jizhong Han, 
	Songlin Hu
}
\email{{zhouwei,hanjizhong,husonglin}@iie.ac.cn}
\affiliation{
  \institution{Institute of Information Engineering, Chinese Academy of Sciences}
}
\authornote{Wei Zhou and Songlin Hu are the corresponding authors.}





\begin{abstract}
Automatically labeling multiple styles for every song is a comprehensive application in all kinds of music websites. Recently, some researches explore review-driven multi-label music style classification and exploit style correlations for this task. However, their methods focus on mining the statistical relations between different music styles and only consider shallow style relations. Moreover, these statistical relations suffer from the underfitting problem because some music styles have little training data. 

To tackle these problems, we propose a novel knowledge relations integrated framework (KRF) to capture the complete style correlations, which jointly exploits the inherent relations between music styles according to external knowledge and their statistical relations. Based on the two types of relations, we use graph convolutional network to learn the deep correlations between styles automatically. Experimental results show that our framework significantly outperforms the state-of-the-art methods. Further studies demonstrate that our framework can effectively alleviate the underfitting problem and learn meaningful style correlations. The source code can be available at https://github.com/Makwen1995/MusicGenre. 
\end{abstract}

%



\maketitle

\section{Introduction}

With the development of multimedia technologies and the growth of available information, music classification has attracted more and more attention in recent years. Specifically, music style classification is a practical and important task since music style is the most common way people use to organize millions of songs, albums, and artists on the Internet. It is often applied to many fields such as music recommendation and music information retrieval.

\begin{figure}[!htbp]
	\centering
	\includegraphics[scale=0.5]{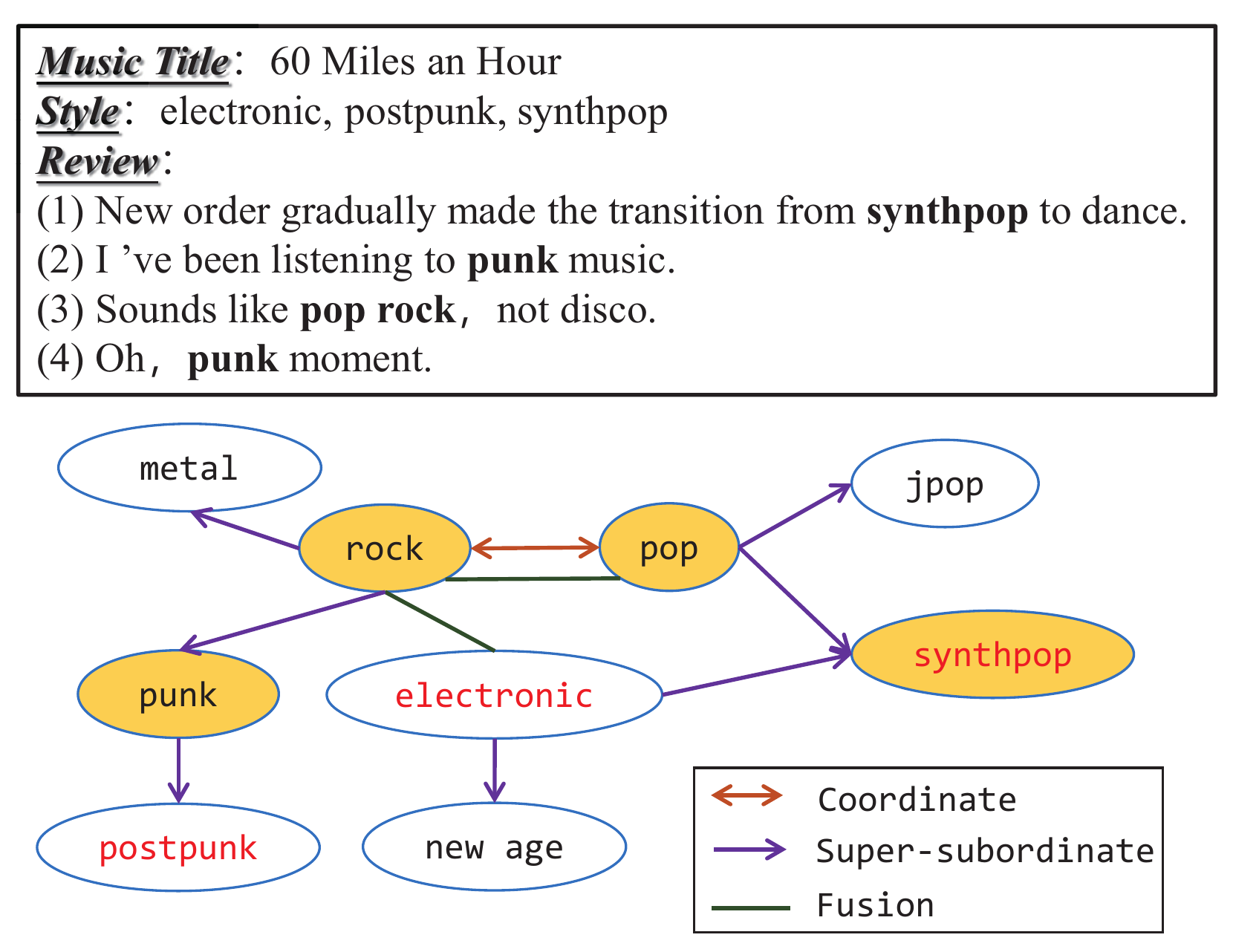}
	\caption{An example of music style correlations. Words indicative of the style in reviews are noted in bold in the text and shaded in yellow in the graph, and the ground truth labels are noted in red in the graph.}
	\label{example}
\end{figure}

Most previous studies traditionally consider music style classification as a single-label classification problem~\cite{nanni2016combining,tsaptsinos2017lyrics,tang2018music}. Nevertheless, a music piece could belong to more than one different style at the same time. For example, a song could be Pop, and has elements from Jazz and Electronic at the same time. Therefore, single-label classification cannot model this kind of multiplicity, and it stimulates research dealing with multi-label music style classification. 

Moreover, it is not only desirable to classify the music with their multiple styles but also to exploit style correlations due to the co-occurrences of the styles of interest. While most prior works typically assume music style labels to be independent and adapt either a multi-class or binary classifier to label music pieces. Different from these studies, Zhao et al.~\cite{zhao2019driven} have firstly explored to utilize the statistical correlations between music styles for multi-label music style classification. Their method has achieved superior performance in the task, which shows it is a promising direction to improve classification performance.

However, the style correlations based on statistics are limited and incomplete. For one thing, the co-occurrence relations cannot reflect the development and evolution relations of the music styles. For another, some categories of music style are very common such as rock or pop music (the majority class), while most categories have many fewer samples (the minority class), forming a long-tail distribution when constructing statistical correlation matrix. This phenomenon may lead to the model overfitting the majority class and underfitting the minority class.

To solve the above problems, we integrate external knowledge into style correlations. We show an example in Figure~\ref{example} to illustrate how external knowledge can help in this task. According to music knowledge, we model style dependencies with three types of style relations in a knowledge graph. For this example, the ground truth labels of `60 Miles an Hour' are \textit{`Electronic'}, \textit{`PostPunk'} and \textit{`SynthPop'}, while only \textit{`SynthPop'} and other styles that are not labels appear in the reviews. However, if these appearing styles are not related to \textit{`PostPunk'} and \textit{`Electronic'} according to statistical data, it is hard to infer correct labels. Therefore, it is natural and necessary to explore inherent relations between styles from knowledge. Intuitively, \textit{`PostPunk'} is what \textit{`Punk'} evolved into (i.e., the super-subordinate relation), and \textit{`Electronic rock'} is a broad music style that involves \textit{`Rock'} and \textit{`Electronic'} music (i.e., the fusion relation). Through these knowledge relations, it is more likely to infer correct labels from the \textit{`Punk'} indicator and \textit{`Rock'} indicator.

In addition to the above problem, the relations between music styles are complicated and interconnected, and there are deep relationships between music styles, such as multi-hop relations. Zhao et al.~\cite{zhao2019driven} only adopted the dot product to measure the proximity of music styles from correlation matrix, and thus they can only preserve the first-order proximity between style labels, which are insufficient to capture the deep relationships between music styles.

In this paper, we investigate: (1) how to overcome the limitation of statistical correlations between music styles; and (2) how to capture the deep correlations between styles according to music styles dependencies. To solve the first challenge, we focus on integrating external knowledge about music styles and statistical information to jointly derive proper and complete dependencies between different styles, which can alleviate the overfitting and underfitting problem. To capture the deep correlations between styles, we employ Graph Convolution Network (GCN) to learn style representations using the above style dependencies to guide the information propagation among nodes (styles) in GCN. Finally, we fuse the review representations learned from hierarchical attention network (HAN) and the style representations for music style classification. We evaluate our framework based on two real-world datasets, and the results show that the proposed framework outperforms all the comparison methods with a large margin. Further studies demonstrate our ability to effectively alleviate the problem of underfitting the minority class and learn meaningful style correlations.

The contributions of this paper can be summarized as follows: 
\begin{itemize}
    \item We integrate external music knowledge and statistical relations between style labels to exploit style correlations.
    \item To capture the deep correlations between styles, we learn the latent style representations with GCN, which is guided by the style correlation matrix we designed.
    \item We fuse review representations and style representations for music style classification. 
\end{itemize}

\section{Related Work}
In this section, we first review the traditional music style classification methods, followed by reviewing the representative studies on multi-label music style classification, which is a more realistic task for music style classification, but more difficult as well.

\subsection{Music Style Classification}
The existing methods for music style classification can be classified into three main categories: (1) audio signal-based,  (2) text-based, and (3) deep learning-based methods.

The audio signal-based approaches rely on feature extraction from audio information. Traditional studies have adopted machine learning methods with hand-crafted audio features, such as Support Vector Machine (SVM)~\cite{xu2003musical}, Hidden Markov Model (HMM)~\cite{chai2001folk,pikrakis2006classification}, and Decision Tree (DT)~\cite{zhou2006music}. 

However, in real-world applications, audio information is not always available due to copyright restrictions, so the methods using audio features have certain limitation and lack generalization ability. Motivated by the fact that audio information could be unavailable or inadequate, several text-based methods have been exploited for this task. Based on music reviews, Hu et al.~\cite{hu2005mining} used a naive bayesian classifier to predict music styles. They extended these experiments with agglomerative hierarchical clustering for predicting usages of music in 2006~\cite{hu2006stylistics}. 

Moreover, Fell et al.~\cite{fell2014lyrics} considered the important contribution of lyrics when analysing and classifying music and proposed a lyrics-based model to classify music by 13 stylistic and linguistic features. Tsaptsinos~\cite{tsaptsinos2017lyrics} explored a hierarchical attention network for style classification of intact lyrics. 
Oramas et al.~\cite{oramas2016exploring} enriched review texts semantically with named entity disambiguation and extracted linguistic, sentimental and semantic features as inputs to an SVM classifier for style classification.

More recently, neural network methods have also been utilized for this task. Some approaches employed visual representations of audio signal in the form of the spectrogram, which were used as inputs to Convolution Neural Networks (CNNs)~\cite{dieleman2014end, pons2016experimenting,choi2016automatic}, following approaches similar to those used for image classification. 
Choi et al.~\cite{choi2017convolutional} proposed a convolutional recurrent neural network through spectrogram for music style classification. 
There were other approaches that do not rely on the spectrogram representation of audio. Tang~\cite{tang2018music} explored a hierarchical long short-term memory (LSTM) model in music style classification. 

\subsection{Multi-Label Classification}
The traditional multi-label classification methods can be classified into two main types: problem transformation methods and algorithm adaptation methods. Problem transformation methods are multi-label learning methods that transform the multi-label problem into multiple single-label problems, so the existing single-label algorithms can be used. Whereas algorithm adaptation methods focus on adapting, extending, and customizing existing machine learning algorithms for the task of multi-label learning. 

Multi-label classification is a widely studied problem in diverse applications such as audio classification~\cite{Boutell_2004} and information retrieval~\cite{Zhu_2005,Gopal_2010}. However, there were few approaches for multi-label classification of music style, Sanden et al.~\cite{Sanden_2011} used ensemble techniques in the task and obtained better results than just using a single classifier. Wang et al.~\cite{wang2009tag} proposed a hyper-graph integrated SVM method, which can integrate both music contents and music tags for automatic music style classification. Oramas et al.~\cite{oramas2018multimodal} proposed an approach to learn and combine multi-modal data representations for music style classification. 


These methods above do not consider the relationships between labels, Zhao et al. \cite{zhao2019driven} firstly explored review-driven multi-label music style classification and exploited the style correlations. However, they only explored shallow style relations based on statistics, which suffer from the underfitting problem.


\section{Problem Statement and Preliminary}
\subsection{Problem Statement}
Given several reviews from a piece of music, this task requires models to predict a set of music styles. Assume that $X=\left\{x_{1},x_{2},\ldots ,x_{K}\right\}$ denotes the input $K$ reviews, and each review $x_i$ contains $J$ words $x_i=x_{i1}, x_{i2},\ldots ,x_{iJ}$. The term $C=\left\{c_{1},c_{2},\ldots ,c_{\left| \mathcal{C}\right|}\right\}$ denotes the set of style categories, which contains $\left| \mathcal{C} \right| $ style categories. Every sample $X$ has a gold set with $M$ style labels, and $M$ varies in different samples. The target of review-based multi-label music style classification is to learn the mapping from input reviews to style labels.

\subsection{Build the Knowledge Graph}\label{build_knowledge_graph}
Before detailing our proposed framework, we briefly introduce how we built the knowledge graph of related music styles in our work. Firstly, we consider Wikipedia\footnote{https://en.wikipedia.org/wiki/List\_of\_popular\_music\_genres} and AllMusic\footnote{https://www.allmusic.com/genres} as the source for constructing the knowledge graph since it is relatively authoritative and contains a hierarchical taxonomy of different music styles.
\begin{figure}[!htbp]
	\centering
 	\includegraphics[scale=0.3]{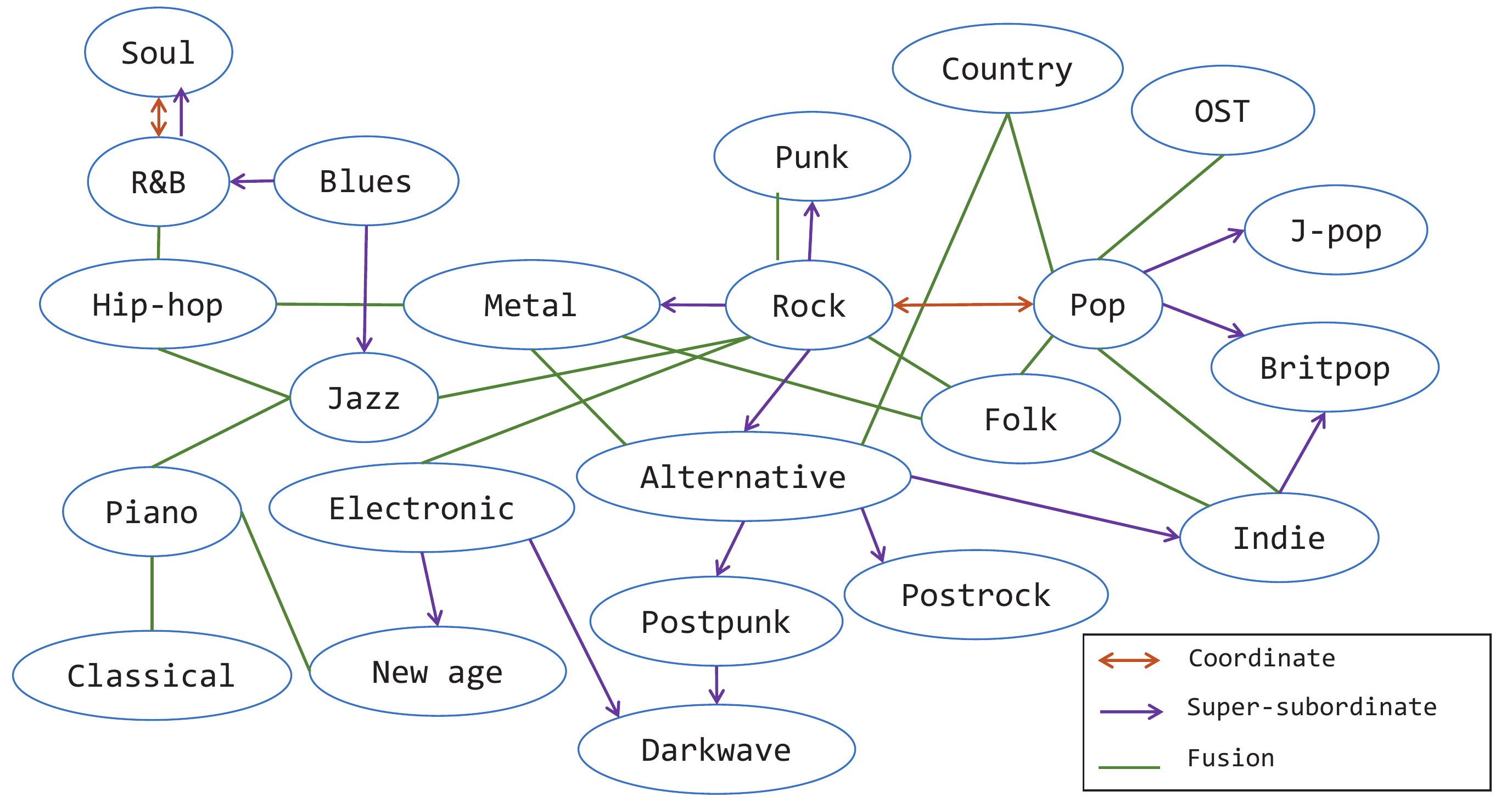}
	\caption{Music sytle relation.}
	\label{knowledge_graph}
\end{figure}
Secondly, We define 3 types of style relations for the knowledge graph: super-subordinate relation, coordinate relation, and fusion relation:
\begin{itemize}
    \item \textbf{Super-subordinate relation}, also called ISA relation, indicates that one style is a subgenre of another style, e.g., \textit{punk music, metal music} and \textit{alternative music} are all subgenres of \textit{rock music}. 
    \item \textbf{Coordinate relation} means that the two styles are very similar, which indicates these two music styles are closely interconnected and mutually inclusive with each other, e.g., \textit{rock music} and \textit{pop music}.
    \item \textbf{Fusion relation} with which style pairs can form a hybrid music style together, e.g., \textit{folk rock music}\footnote{https://en.wikipedia.org/wiki/Folk\_rock} is a hybrid music style combining elements of \textit{folk music} and \textit{rock music}.
\end{itemize}


\section{Model}
\begin{figure*}[htbp]
	\centering
	\includegraphics[scale=0.6]{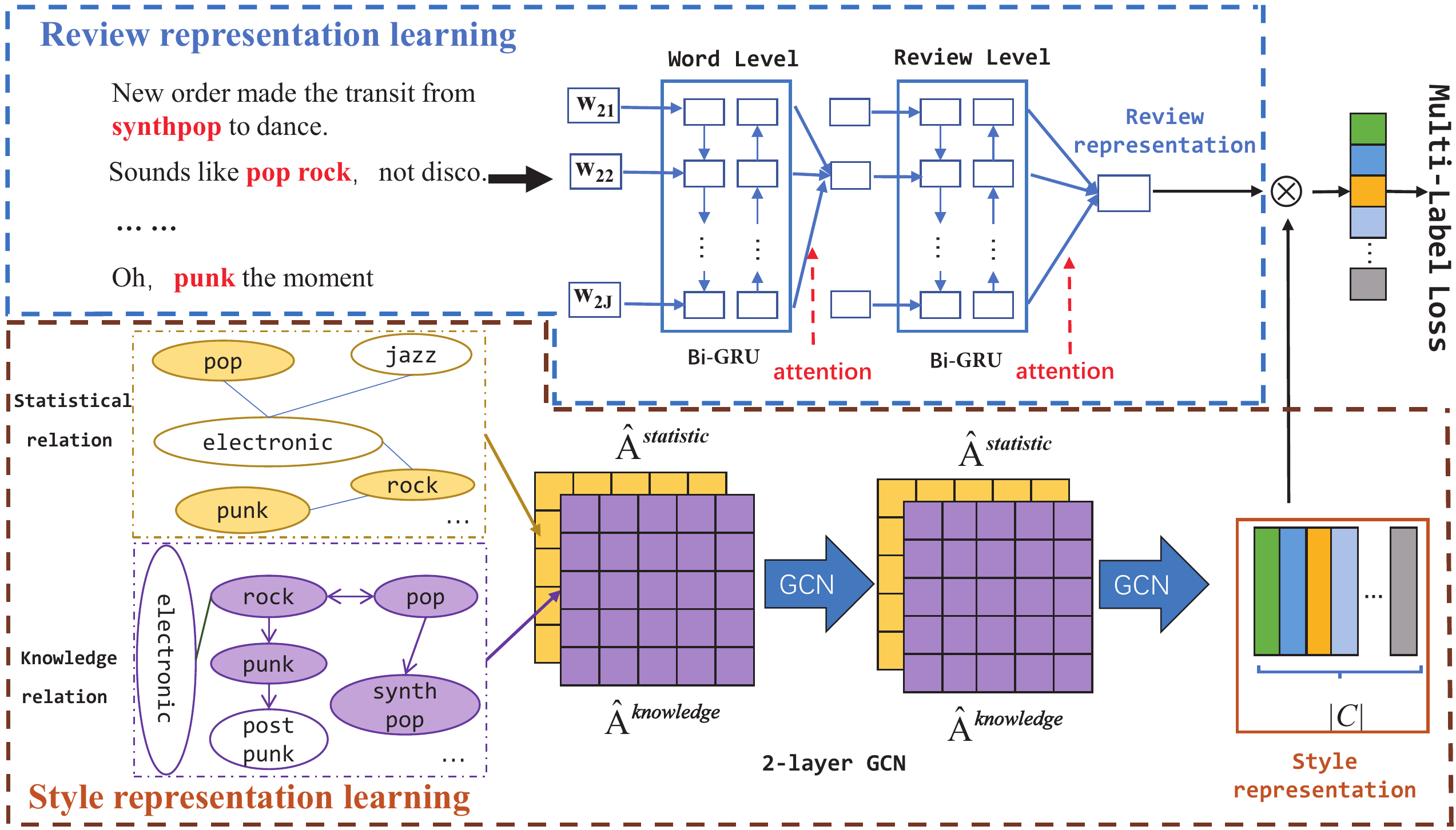}
	\caption{An illustration of the knowledge relations integrated framework KRF. $|C|$ is the number of style categories.}
	\label{model_structure}
\end{figure*}

An illustration of proposed framework is shown in Figure~\ref{model_structure}, which is composed of two major components: review representation learning and style representation learning base on style correlations. Specifically, the review representation learning describes the mapping of review sequence from word embedding to semantic space; and the style representation learning illustrates how to integrate the style correlations from data and external knowledge, and encode them into the style label representation. Next, we describe the details of different components in the following sections.

\subsection{Review Representation Learning}
To obtain the representation of reviews from different semantic levels, we model the review sequence via a hierarchical attention network (HAN)~\cite{yang2016hierarchical}. To let the framework pay more or less attention to individual words and reviews, this module includes two levels of attention mechanisms: word level part containing a bidirectional Gated Recurrent Unit (GRU) layer and an attention layer, review level part containing a bidirectional GRU layer and an attention layer. 

\textbf{Word Level}: 
Given a review text with $J$ words, we first embed each word in the review into a word vector $\mathbf{x_{ij}} \in \mathbb{R}^{d}$, where $d$ is the dimensionality of word embedding vector. 
Then, we employ a bidirectional GRU to capture contextual information from two directions of the word sequence, the forward GRU cell generates a hidden state $\mathbf{\overrightarrow {h_{ij}}}$ and the backward GRU cell generates a hidden state $\mathbf{\overleftarrow {h_{ij}}}$. We concatenate two hidden states to obtain the final state: 
\begin{equation}
\begin{split}
& \overrightarrow {h}_{ij} = \overrightarrow{GRU}\left ( x_{ij} \right ), j\in \left[ 1,J\right]   \,, \\
& \overleftarrow {h}_{ij} = \overleftarrow{GRU}\left ( x_{ij} \right ), j\in \left[ J,1\right]   \,, \\
& h_{ij} = \overrightarrow {h}_{ij} \oplus \overleftarrow {h}_{ij} \,, \\
\end{split}
\end{equation}
where $\oplus$ denotes the concatenation operation.

Obviously, not all words contribute equally to the representation of the review meaning, some words are more important than others. Hence, we use attention mechanism to weigh the importance of every word and obtain an enhanced review representation, which can be formulated as:
\begin{equation}
\begin{split}
& \mathbf{u}_{ij} = \tanh \left( \mathbf{W}_{w}\mathbf{h}_{ij} + \mathbf{b}_w \right) \,, \\
& \alpha_{ij} =\frac{\exp \left( u_{ij}^{T} u_w \right) }{\sum_j \exp \left( u_{ij}^{T} u_w \right) } \,, \\
& \mathbf{x}_i     = \sum_{j} \alpha_{ij} h_{ij} \,,  \\
\end{split}
\end{equation}
where $\mathbf{W}_{w}$ and $\mathbf{b}_w$ denote the weight matrix and the bias term, then we measure the importance of the word by computing the similarity between $\mathbf{u}_{ij}$ and a word-level context vector $\mathbf{u}_W$. After obtaining a normalized importance weight $\alpha _{ij}$ through a softmax function, we compute the review vector $\mathbf{x}_i$ as a weighted sum of the word annotations based on the weights.

\textbf{Review Level}: 
After getting the review vector $\mathbf{x}_i$, we use Bi-GRU and attention mechanism to encode the review sequence $\left\{x_{1},x_{2},\ldots ,x_{K}\right\}$ in a similar way. Finally, we can obtain the review sequence vector $\mathbf{X} \in \mathbb{R}^{d}$.

\subsection{Style Representation Learning}
To incorporate the style correlations from statistical information and external knowledge in style representation, we first build a correlation matrix with fusing the statistical and knowledge relations between styles. That is, we define style correlations not only via mining their co-occurrence patterns within the dataset, but also via exploiting external knowledge about music styles.

Then, we utilize GCN~\cite{kipf2017semi} to learn the deep relationships between styles. GCN is designed for learn feature representations for nodes based on propagating messages between neighboring nodes, and an adjacency matrix is used to represent the graph structure information. Thus, the adjacency matrix is critical for guiding information propagation in GCN. Inspired by~\cite{Chen_2019_CVPR}, we apply the above integrated correlation matrix to GCN instead of adjacency matrix.

\subsubsection{\textbf{Statistical Correlation Matrix}}

To capture the co-occurrence patteran of styles, we count the number of occurrences of style pairs in the training set and get the matrix $\mathbf{A} \in \mathbb{R}^{|C| \times |C|}$. Concretely, $\mathcal{|C|}$ is the number of style categories, and $A_{ij}$ denotes the co-occurrence times of $c_i$ and $c_j$. 

However, there are two drawbacks in this style co-occurrence matrix. Firstly, the co-occurrence patterns between style pairs may form a long-tail distribution, where some rare co-occurrences may be noise. Secondly, the number of co-occurrences from a dataset cannot reflect the style co-occurrences in reality well. 

Thus, we set a threshold $\tau$ to filter out noisy edges:
\begin{equation}\label{data_correlation_matrix}
    \mathbf{A}^{statistic}_{ij}=\begin{cases}0, \quad & \text{if $a_{ij} < \tau$} \\
            A_{ij}, & \text{otherwise} \end{cases} \,.
\end{equation}
Then, we can get the filtered correlation matrix $\mathbf{A}^{statistic}$, which exploits the statistical relations between styles. 
\subsubsection{\textbf{Knowledge Correlation Matrix}}
Because the co-occurrence matrix cannot capture the inherent relations between style labels in reality, we leverage a knowledge graph $\mathcal{G}$ that captures the structural relationships of labels. Formally, we define the knowledge graph $\mathcal{G}=(\mathcal {E},\mathcal {R}) $, where $\mathcal{E}$ and $\mathcal{R}$ are the set of entities and relations in the knowledge graph, respectively. In this task, a label $c \in C$ corresponds to an entity $e \in \mathcal{E}$, and the relation existing between labels corresponds to a relation $r \in \mathcal{R}$. The relation set $\mathcal{R}$ contains three types of relations, super-subordinate relations $r_{super-subordinate}$ , coordinate relations $r_{coordinate}$ and fusion relations $r_{fusion}$, which are defined in Section~\ref{build_knowledge_graph}.

Then, we use a scoring function $\mathbf{s_r}$ to compute the score between two labels, where $\mathbf{s_r}$ characterizes the importance of relation $r$ between labels and depends on the type of relation $r$, and the knowledge correlation matrix is defined as follows:
\begin{equation}
    \mathbf{A}^{knowledge}_{ij}= \begin{cases} s_r \left ( e_i , e_j \right ) ,\quad & \text{if $r \in \mathcal{R} $} \\
            0, & \text{otherwise} \end{cases} \,,
\end{equation}
where $\mathbf{A}^{knowledge}_{ij} \in \mathbb{R}^{|C| \times |C|}$.

For simplicity, we set $\mathbf{s_{r}\left( \cdot \right)}$ as a constant function: 1 when the relation $r \in r_{fusion}$, 2 when $r \in r_{super-subordinate}$, and 3 when $r \in r_{coordinate}$. 
\subsubsection{\textbf{Integrated Correlation Matrix}} \label{correlation_matrix}
The two correlation matrices obtained from different scales cannot be merged directly, so it is necessary to normalize $\mathbf{A}^{statistic}$ and $\mathbf{A}^{knowledge}$ respectively. The normalization method~\cite{kipf2017semi} is formalized as follows:
\begin{align}
\widehat {\mathbf{A}} &= \mathbf{D}^{-\frac{1}{2}}\mathbf{AD}^{-\frac{1}{2}} \,,
\end{align}
where $\mathbf{D}$ is a diagonal degree matrix with entries $D_{ij} = \Sigma _{j}A_{ij}$

Next, we merge the two correlation matrices together after normalization: 
\begin{align}
\mathbf{A}^{integrated} = [\widehat {\mathbf{A}}^{statistic}; \widehat {\mathbf{A}}^{knowledge}] \,.
\end{align}
Finally, we get the correlation matrix $\mathbf{A}^{integrated} \in \mathbb{R}^{ 2 \times |C| \times |C|}$ in which integrates statistical relations and knowledge relations.

\subsubsection{\textbf{GCN}}
GCNs are neural networks operating on graphs and generating node representations by aggregating neighborhood information. The key idea is to update the feature representations of nodes by propagating information among them. The layer-wise propagation rule of a multi-layer GCN is as follows:
\begin{equation}
\mathbf{H}^{l+1} = \sigma\left( \mathbf{A}^{integrated}\mathbf{H}^{l}\mathbf{W}^{l}\right) \,,
\end{equation}
where $\sigma\left( \cdot \right)$ denotes LeakyRELU~\cite{maas2013rectifier} activation function. $\mathbf{H}^{l} \in \mathbb{R}^{|C| \times D}$ is the matrix of activations in the $l^{th}$ layer (where $|C|$ denotes the number of nodes, $D$ is the dimensionality of node features) and $\mathbf{H}^{l+1}$ is the updated node features. $\mathbf{W}^l \in \mathbb{R}^{D \times D'}$ is a transformation matrix to be learned.

Each GCN layer ${l}$ takes the label representations from previous layer $\mathbf{H}^l$ as inputs and outputs new label representations, i.e., $\mathbf{H}^{l+1}$. 
Depending on how many convolutional layers are used, GCN can capture information only about immediate neighbors (with one convolutional layer) or any nodes at most K-hops neighbors (if K layers are stacked on top of each other). See~\cite{kipf2017semi} for more details about GCN.

We use a 2-layer GCN to learn the deep correlations between music styles. The first layer takes a randomly initialized label embeddings $\mathbf{H^0} \in \mathbb{R}^{|C| \times D}$ as inputs, where $D$ is the dimensionality of the initial label embeddings); and the last layer outputs $\mathbf{H^2} \in R^{|C| \times D'}$ with $D'$ denoting the dimensionality of the final label representations.

\subsection{Music Style Classification}
After the above procedures, we have obtained the review representations $\mathbf{X}$ and the label representations(i.e., $\mathbf{H^2}$). Both representations are important when predict music styles, and thus we employ matrix multiplication to fuse them together.

We apply a nonlinear function ReLU to review representations before the fusion operation, and the final predicted scores $\widehat {y}$ can be obtained as follows:
\begin{align}
\mathbf{X'} &= ReLU(\mathbf{X}) \,, \\
\widehat {y} &=\mathbf{X'W} \left( \mathbf{H^2} \right)^{T} \,,
\end{align}
where $\mathbf{W}^l \in \mathbb{R}^{d \times D'}$ is a linear transformation matrix, and $\mathbf{H^2} \in R^{|C| \times D'}$ is the final label representations learned by the 2-layer GCN.

We use $y \in \mathcal{R}^{|C|}$ to represent the ground-truth label of a music piece, where $y_i = {0, 1}$ denotes whether label i appears in the music or not. The whole framework is trained with the multi-label cross entropy loss which can be formulated as follows:
\begin{align}
\mathcal{L} &= \sum ^{C}_{c=1}y^{c}\log \left( \sigma \left( {\widehat {y}}^c \right) \right) + \left( 1-y^c \right) \log \left( 1 - \sigma \left( \widehat {y}^c \right) \right) \,,
\end{align}
where $\sigma$ is the sigmoid function.

\section{Experiment}
\subsection{Dataset}
We evaluate the proposed framework on two real-word datasets: Douban Music~\cite{zhao2019driven} and Amazon Music, which were originally collected from the most popular music review website in China\footnote{https://music.douban.com} and the U.S.\footnote{https://www.amazon.com} respectively. The detailed statistics of these two datasets are listed in Table~\ref{tab:datset}.

\begin{table}[!htbp]
    \centering
    \caption{Dataset statistics.}
	\setlength{\tabcolsep}{4mm}{
        \begin{tabular}{c|c c}
            \toprule
            \textbf{Statistic}&\textbf{Douban} &\textbf{Amazon} \\
            \hline
            \hline
            \ Songs& 7,172 & 9,981 \\
            \hline
            \ Reviews& 293,426 & 275,922  \\
            \hline
            \ Styles& 22 & 20 \\
            \hline
            \ Avg. \# of tags& 2.2 & 2.7 \\
            \hline
            \ Min. \# of tags& 2 & 1 \\
            \hline
            \ Max. \# of tags& 5 & 13 \\
            \bottomrule
        \end{tabular}
    }
    \label{tab:datset}
\end{table}

\noindent\textbf{Douban Music~\cite{zhao2019driven}}: The dataset contains over 7.1k samples, 288K reviews, and 22 music styles. Each sample includes a music title, a set of human annotated styles, and top 40 voted reviews. 

\noindent\textbf{Amazon Music}: In order to verify the effectiveness and generality of our framework, we also model on the method used in~\cite{zhao2019driven} to crawl music data with user reviews from Amazon. Amazon has its own hierarchical taxonomy of music styles, which is up to 500 styles overall. In our dataset, we keep the 20 styles that satisfy the condition of having been annotated in Douban Music. The dataset contains over 9k samples, 276k reviews, and 20 music styles.


\begin{table*}[!htbp]
  \caption{Comparisons with state-of-the-art methods on the music review datasets. ``+'' represents that higher scores are better and ``-'' represents that lower scores are better. It can be seen that the proposed framework significantly outperforms the baselines. The experimental results of baselines are directly cited from~\cite{zhao2019driven}.}

  \label{tab:experiment}
  \setlength{\tabcolsep}{2.5mm}{
      \begin{tabular}{c|cccc|cccc}
                \toprule
                \multirow{2}{*}{\textbf{Models}} & \multicolumn{4}{c|}{\textbf{Douban music}} & \multicolumn{4}{c}{\textbf{Amazon music}} \\
                &\textbf{OE (-)} &\textbf{HL (-)} &\textbf{Macro F1 (+)} &\textbf{Micro F1 (+)}
                &\textbf{OE (-)} &\textbf{HL (-)} &\textbf{Macro F1 (+)} &\textbf{Micro F1 (+)}
                \\
                \midrule
                \midrule
                ML-KNN~\cite{zhang2007ml}                   & 77.3 & 0.094 & 23.6 & 38.1    &  55.4&  0.131& 34.1& 47.1 \\
                Binary Relevance~\cite{tsoumakas2009mining} & 74.4 & 0.083 & 24.7 & 41.8    & 41.3 & 0.125 & 35.6 & 48.4 \\
                Classifier Chains~\cite{read2011classifier} & 67.5 & 0.107 & 29.9 & 44.3    & 42.5 & 0.132 & 36.9 & 50.4 \\
                Label Powerset~\cite{tsoumakas2007random}   & 56.2 & 0.096 & 37.7 & 50.3    & 28.1 & 0.114 & 40.2 & 55.7 \\
                \midrule
                MLP~\cite{zhao2019driven}       & 71.5 & 0.081 & 29.8 & 45.8   & 24.5 & 0.104 & 19.0 & 46.3 \\
                CNN~\cite{zhao2019driven}       & 37.9 & 0.099 & 32.5 & 49.3   & 23.7 & 0.098 & 42.0 & 58.9 \\
                LSTM~\cite{zhao2019driven}      & 30.5 & 0.089 & 33.0 & 53.9   & 23.7 & 0.101 & 34.5 & 54.6  \\
                HAN-LCM~\cite{zhao2019driven}  & 22.6 & 0.074 & 54.4 & 64.5   & 17.8 & 0.086 & 56.2 & 69.1  \\
                \midrule
                \textbf{KRF} & \textbf{15.5} & \textbf{0.058}  & \textbf{66.4} & \textbf{70.8}
                & \textbf{16.6} & \textbf{0.070} & \textbf{60.8} & \textbf{73.2} \\
                \bottomrule
        \end{tabular}
    }
\end{table*}

\subsection{Baseline Models}
We compare KRF with several baseline methods, including widely-used multi-label classification methods and neural network methods such as LSTM, CNN, and HAN-LCM. These baseline methods and results are reported in ~\cite{zhao2019driven}.
\begin{itemize}
    \item ML-KNN: It is a multi-label learning algorithm developed on the basis of K-Nearest Neighbor (KNN) algorithm.
    \item Binary Relevance: It is a problem transformation algorithm that decomposes the multi-label classification problem into a number of independent binary classification problems without considering the dependencies among labels.
    \item Classifier Chains: It transforms the multi-label learning problem into a chain of multiple binary classification problems and takes label dependencies into account.
    \item Label Powerset: It transforms the multi-label learning problem into one multi-class single label learning problem.
    \item MLP: It takes the music representations as inputs to a multi-layer perceptron, and generates the probability of music styles through a sigmoid layer.
    \item CNN: It contains two layers of CNN, which have multiple convolution kernels.
    \item LSTM: It contains two layers of LSTM, which processes words and sentences separately to get the music representations.
    \item HAN-LCM~\cite{zhao2019driven}: It contains a label-graph based neural network and a soft training mechanism with correlation-based label representation. It is the state-of-the-art model until this publication.
\end{itemize}

\subsection{Evaluation Metric and Experiment Settings}
Multi-label classification requires different evaluation metrics compared with traditional single-label classification. Following the settings of ~\cite{zhao2019driven}, we report the F1 score, one-error, and hamming loss for performance evaluation. The definition of these metrics are listed as follows:
\begin{itemize}
    \item {\verb|F1 score|}: We calculate macro F1 and micro F1, respectively. Macro F1 computes the F1 score independently for each label and then takes the average as the final score. Micro F1 aggregates the contributions of all labels to compute the average F1 score.
    \item {\verb|One-error|}: One-error evaluates the fraction of examples whose top-ranked label is not in the gold label set.
    \item {\verb|Hamming Loss|}: Hamming loss evaluates the fraction of misclassified instance-label pairs, i.e., label not belonging to the instance is predicted or a label belonging to the instance is not predicted. The smaller value of hamming loss means the better performance.
\end{itemize}

Following the settings in~\cite{zhao2019driven}, we randomly (same random seed as~\cite{zhao2019driven}) select 70\% instances as the training dataset, and split the rest with ratio 70:30 for validating and testing respectively. 
Our style correlation learning module consists of two GCN layers with output dimensionality of 512 and 128, respectively. The word embeddings in the framework are initialized with the 128-dimensional word vectors, which are trained on the datasets by Skip-gram~\cite{mikolov2013distributed} algorithm.
We use the Adam optimizer~\cite{kingma2014adam} with the learning rate initialized to 0.001. The maximum training epoch is set to 20, and the batch size of the training set is set to 64. 
We select the best parameter configuration based on performance on the validation set and evaluate the configuration on the test set. 
We implement the network based on PyTorch.\footnote{https://pytorch.org/}

\begin{table*}[htbp]
  \centering
  \caption{Ablation study of our framework KRF.}
  \label{tab:ablation}
  \setlength{\tabcolsep}{2.5mm}{
  \begin{tabular}{c|cccc|cccc}
            \toprule
            \multirow{2}{*}{\textbf{Models}} & \multicolumn{4}{c|}{\textbf{Douban music}} & \multicolumn{4}{c}{\textbf{Amazon music}} \\
            &\textbf{OE (-)} &\textbf{HL (-)} &\textbf{Macro F1 (+)} &\textbf{Micro F1 (+)}
            &\textbf{OE (-)} &\textbf{HL (-)} &\textbf{Macro F1 (+)} &\textbf{Micro F1 (+)}
            \\
            \hline
             \textbf{KRF} & \textbf{15.5} & \textbf{0.058}  & \textbf{66.4} & \textbf{70.8}
                & \textbf{16.6} & \textbf{0.070} & \textbf{60.8} & \textbf{73.2} \\
            \ w/o SRelation & 17.3 & 0.063 & 58.9  & 68.6  & 17.3 & 0.071 & 59.2  & 71.4 \\
            \ w/o KRelation & 19.5 & 0.067 & 45.5 & 64.4  &17.5 & 0.078 & 51.4  & 69.6  \\
            \ Only HAN  & 27.6 & 0.077 & 39.2 & 57.4 
            & 18.2 & 0.084 & 54.9  & 67.6 \\
            \bottomrule
    \end{tabular}}
\end{table*}

\subsection{Experimental Results and Analysis}
Table~\ref{tab:experiment} demonstrates the performance of all the compared methods based on the two datasets. The results indicate that the proposed framework outperforms all the baselines, which confirms the effectiveness of label representations learning with style correlations, which include statistical relations and knowledge relations.

It is observed that the performance of algorithm adaptation methods (\textbf{ML-KNN}) is obviously worse than all the 3 problem transformation methods (\textbf{Binary Relevance, Classifier Chains}, and \textbf{Label Powerset}). This is due to its ignorance of the inter relationship between labels. Furthermore, \textbf{ML-KNN} determines the label set of a new instance by employing the maximum a posteriori principle, so it may be affected by the facts that the instances with and without a particular label are typically highly imbalanced.

Among the three problem transformation methods, \textbf{Label Powerset} performs best because it does take possible correlations between class labels into account, which considers each member of the power set of labels in the training set as a single label. As expected, \textbf{CNN} and \textbf{LSTM} outperform the traditional multi-label classification methods. This observation indicates that the neural network models can learn deep latent features of reviews indicative of styles automatically.

\textbf{HAN-LCM} performs the best among all the baselines, which is not surprising since it exploits the style correlations using a label graph and takes advantage of the attention mechanism to effectively capture discriminative features.

In conclusion, the proposed framework \textbf{KRF} outperforms neural network-based models and widely-used multi-label classification methods, and the one-error and F1 score of music style classification obtain significant improvement. Specifically, on the Douban dataset, \textbf{KRF} increases the micro F1 score of the \textbf{HAN-LCM} method (the best baseline) from 64.5\% to 70.8\%, and reduces one-error from 22.6\% to 15.5\%. On the Amazon dataset, the micro F1 score is boosted from 69.1\% to 73.2\%, and reduces one-error from 17.8\% to 16.6\% respectively. The improvements of the proposed framework demonstrate that the style correlations learning with integrated statistical relations and knowledge relations are generally helpful and effective, and \textbf{KRF} can capture the deeper correlations between style labels than \textbf{HAN-LCM}.

\subsection{Ablation Study}
We perform a series of ablation experiments to examine the relative contributions of statistical relations and knowledge relations. To this end, \textbf{KRF} is compared with its three variants below: 
\begin{itemize}
    \item \textbf{w/o SRelation}: Style representation learning without statistical relations between labels (i.e., only with knowledge relations between labels).
    \item \textbf{w/o KRelation}: Style representation learning without knowledge relations between labels (i.e., only with statistical relations between labels).
    \item \textbf{Only HAN}: Removing style representation learning module and only use the hierarchical attention network for review representation.
\end{itemize}

From experimental results in Table~\ref{tab:ablation}, we can observe that:

(1) Compared with \textbf{HAN-LCM}, the proposed \textbf{KRF} only with statistical style correlations performs better, which demonstrates that \textbf{KRF} can capture the deeper correlations between labels.

(2) Style representation learning with knowledge relations performs better than solely using statistical relations, indicating the importance of knowledge relations. This demonstrates that the model with knowledge relations is insensitive to data and has better stability and robustness, which also shows that the model with knowledge relations can capture label dependencies more effectively and accurately.

(3) The results also show that exploiting both knowledge relations and statistical relations to learn style representations can correctly capture the effective correlations between labels, respectively. However, it is necessary to explore their joint effects to further boost the performance.

\subsection{Visualization Analysis}
\begin{figure}[!htbp]
	\centering
	\includegraphics[scale=0.4]{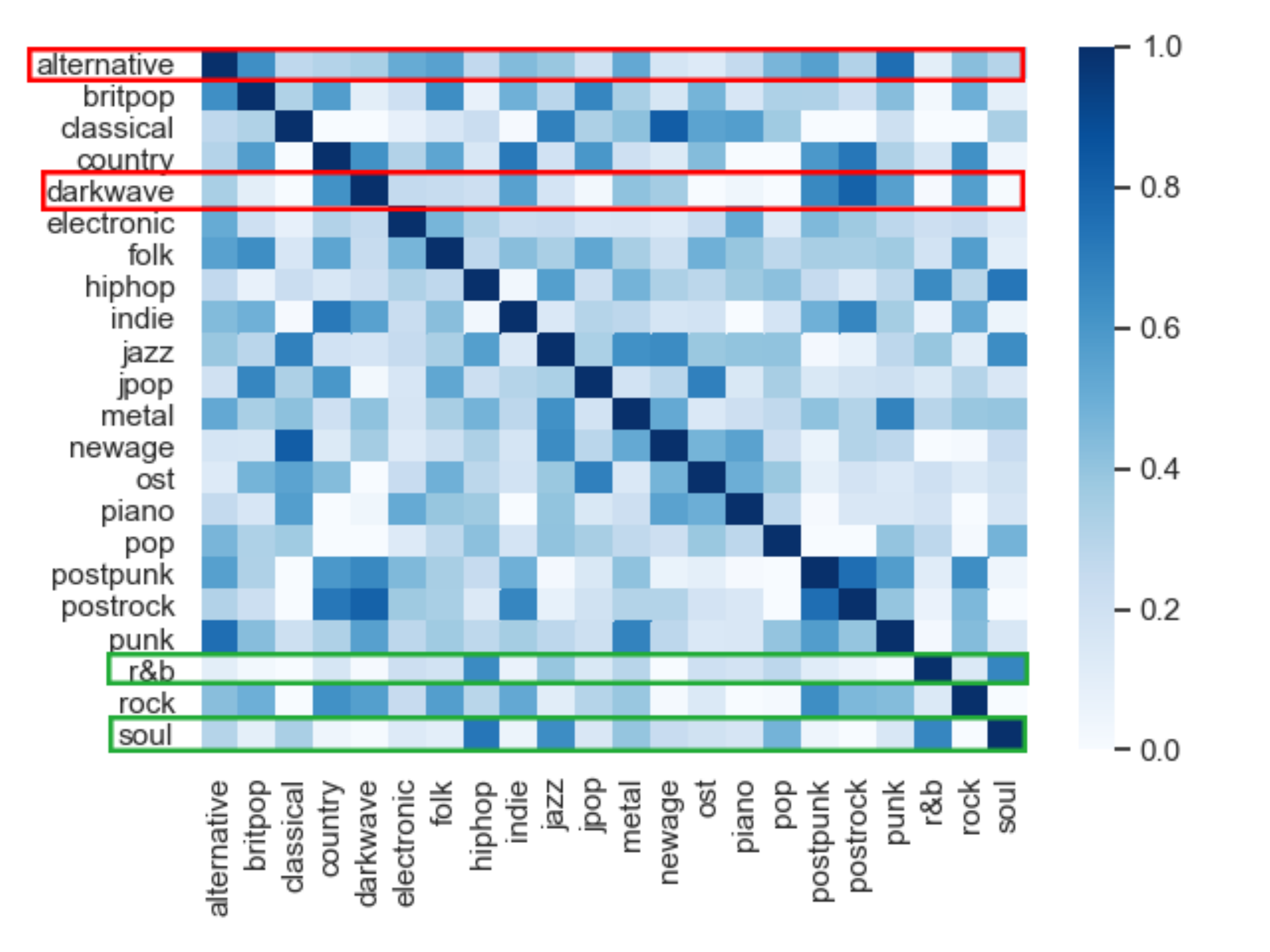} 
	\caption{A heatmap of the automatically learned style label correlations. Darker color represents closer relation.}
	\label{visualization_analysis}
\end{figure}

To gain a clearer view of what is learned by our style representation learning module, we display the heatmap in Figure~\ref{visualization_analysis} to visualize the similarity between style labels. The similarity using in the experiment is computed as a dot product between vectors representing each style label, and the similarity score is scaled to 0 and 1.

We can observe from the figure that the deeper and wider relations between music styles are well captured by GCN, which integrate knowledge relations and statistical relations. 
As shown in the figure, \textit{`Alternative'} is highly linked with three different style labels, \textit{`Britpop'}, \textit{`Punk'} and \textit{`Metal'}. In fact, these 4 styles are all subgenres of \textit{`Rock'} music, specifically, \textit{`Britpop'} music also belongs to \textit{`Alternative'} music. This demonstrates that the proposed \textbf{KRF} can capture the deep label correlations well. 
For \textit{`Dark Wave'}, the most related label is \textit{`Post Rock'}, followed by \textit{`Post Punk'}. Actually, \textit{`Dark Wave'} is derived from \textit{`Post Punk'}, which demonstrates that \textbf{KRF} correctly captures the super-subordinate relations between styles. As for why \textit{`Post Rock'} is more similar to \textit{`Dark Wave'} than \textit{`Post Punk'}, a possible reason is that the proposed \textbf{KRF} comprehensively considers the two types of relationships. 
Moreover, the \textit{`Soul'} and \textit{`R\&B'} show the closest similarity in the figure, and they are both highly linked with two labels, \textit{`Hip hop'} and \textit{`Jazz'}. In reality, these two music styles are highly similar, and the boundary between them is not well-defined, \textbf{KRF} also correctly captures such similar relations.

\subsection{Case Study}
\begin{figure}[!htbp]
	\centering
	\subfigure[Douban Majority Styles]{
		\label{fig:subfig:a1} 
		\includegraphics[width=4.1cm,height=3.5cm]{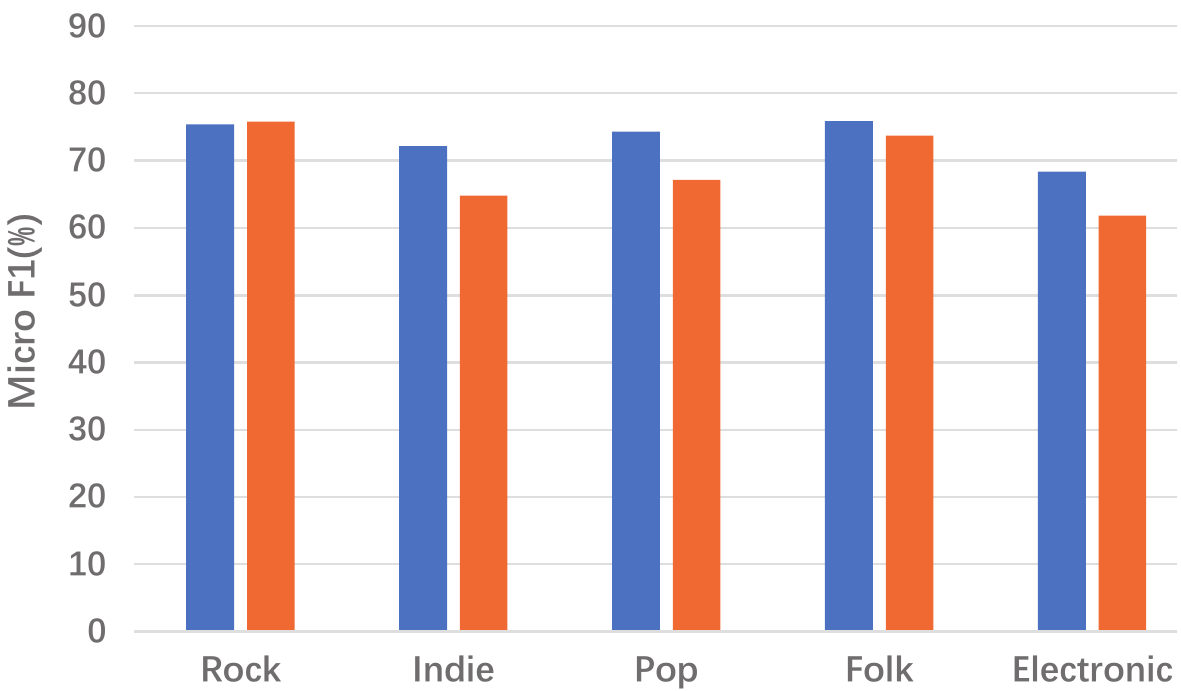}
	}
	\subfigure[Amazon Majority Styles]{
		\label{fig:subfig:a2} 
		\includegraphics[width=3.9cm,height=3.5cm]{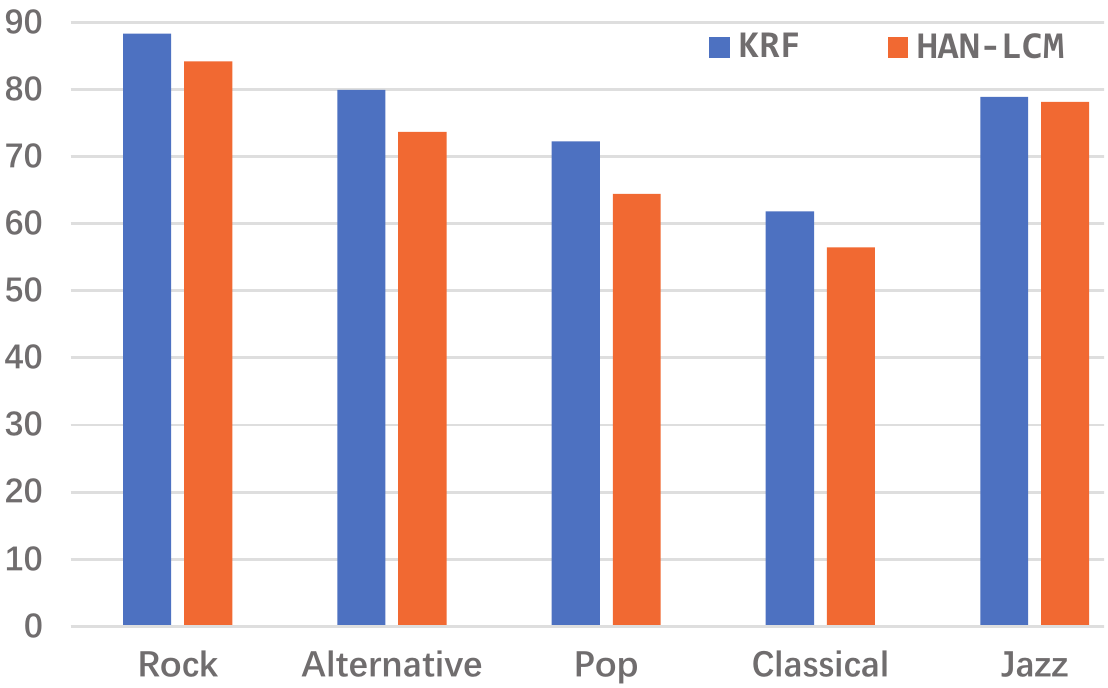}
	}

	\subfigure[Douban Minority Styles]{
		\label{fig:subfig:b1} 
		\includegraphics[width=4cm,height=3.5cm]{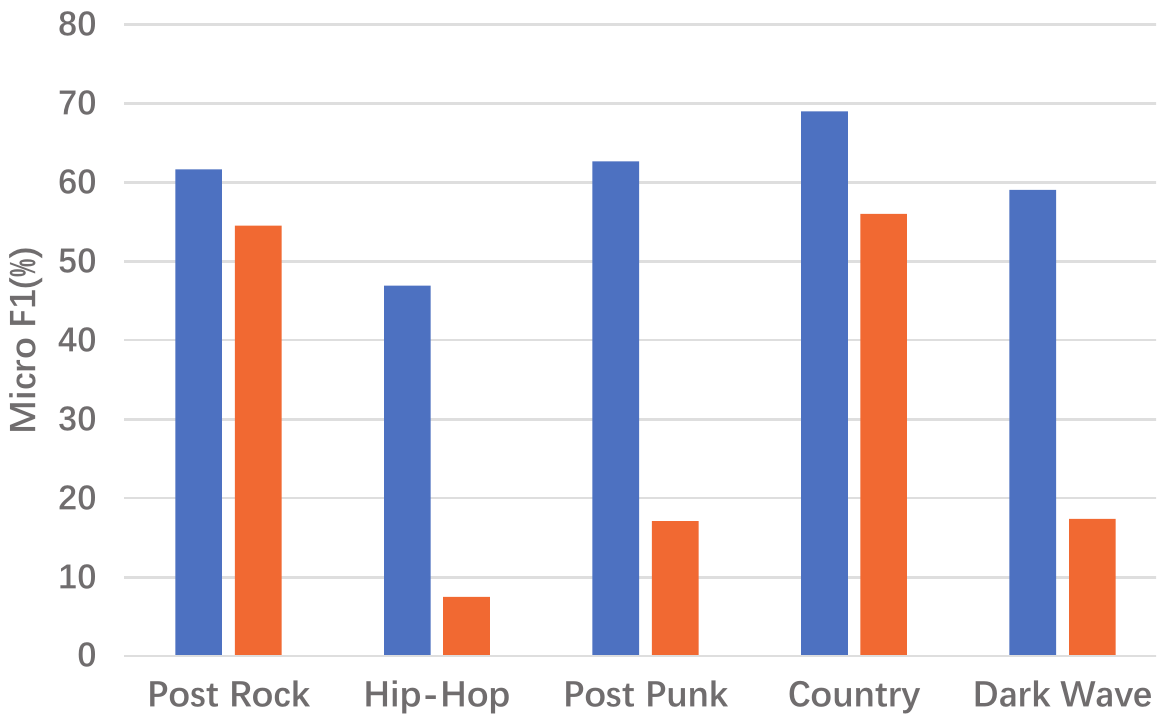} 
	}
	\subfigure[Amazon Minority Styles]{
		\label{fig:subfig:b2} 
		\includegraphics[width=4cm,height=3.5cm]{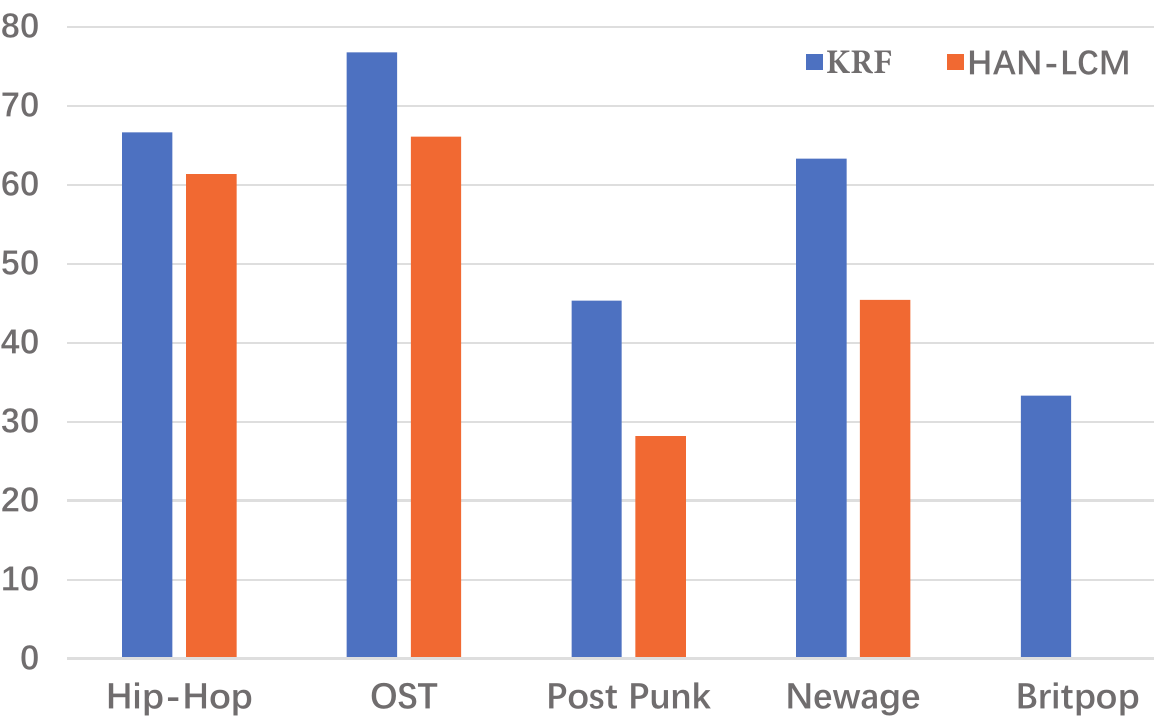} 
	}
	\caption{The performance of the proposed framework on the majority and minority styles. }
	\label{error_analysis}
\end{figure}

Most existing methods~\cite{zhao2019driven} perform worse on the categories with low frequency in the training set. This is because the label distribution is highly imbalanced, where unpopular music styles have too little training data. 
In order to prove the effectiveness of the proposed \textbf{KRF} in alleviating the underfitting problem, we compare the performance on the top 5 majority and minority music styles in Figure~\ref{error_analysis}. 

As shown in Figure~\ref{fig:subfig:a1} and ~\ref{fig:subfig:a2}, the proposed \textbf{KRF} performs better on the top 5 majority music styles than the \textbf{HAN-LCM} model (the best baseline) on two datasets.

Figure~\ref{fig:subfig:b1} and ~\ref{fig:subfig:b2} demonstrate the performance on the top 5 minority music styles. We can observe that \textbf{KRF} has a slight drop, but it still works better than \textbf{HAN-LCM}. Specifically, on the Douban dataset, \textbf{KRF} achieves 7.12\% to 45.55\% absolute improvement on Micro F1 compared with \textbf{HAN-LCM}. On the Amazon dataset, the Micro F1 increased by at least 5.28\% up to 33.33\%. This is because \textbf{KRF} introduces external knowledge about music styles and automatically learns style representations with integrated style correlations. Therefore, the results show that the proposed \textbf{KRF} can effectively alleviate the problem of underfitting rare categories.

\subsection{Parameter Sensitivity}
\begin{figure}[!htbp]
	\centering
	\includegraphics[scale=0.6]{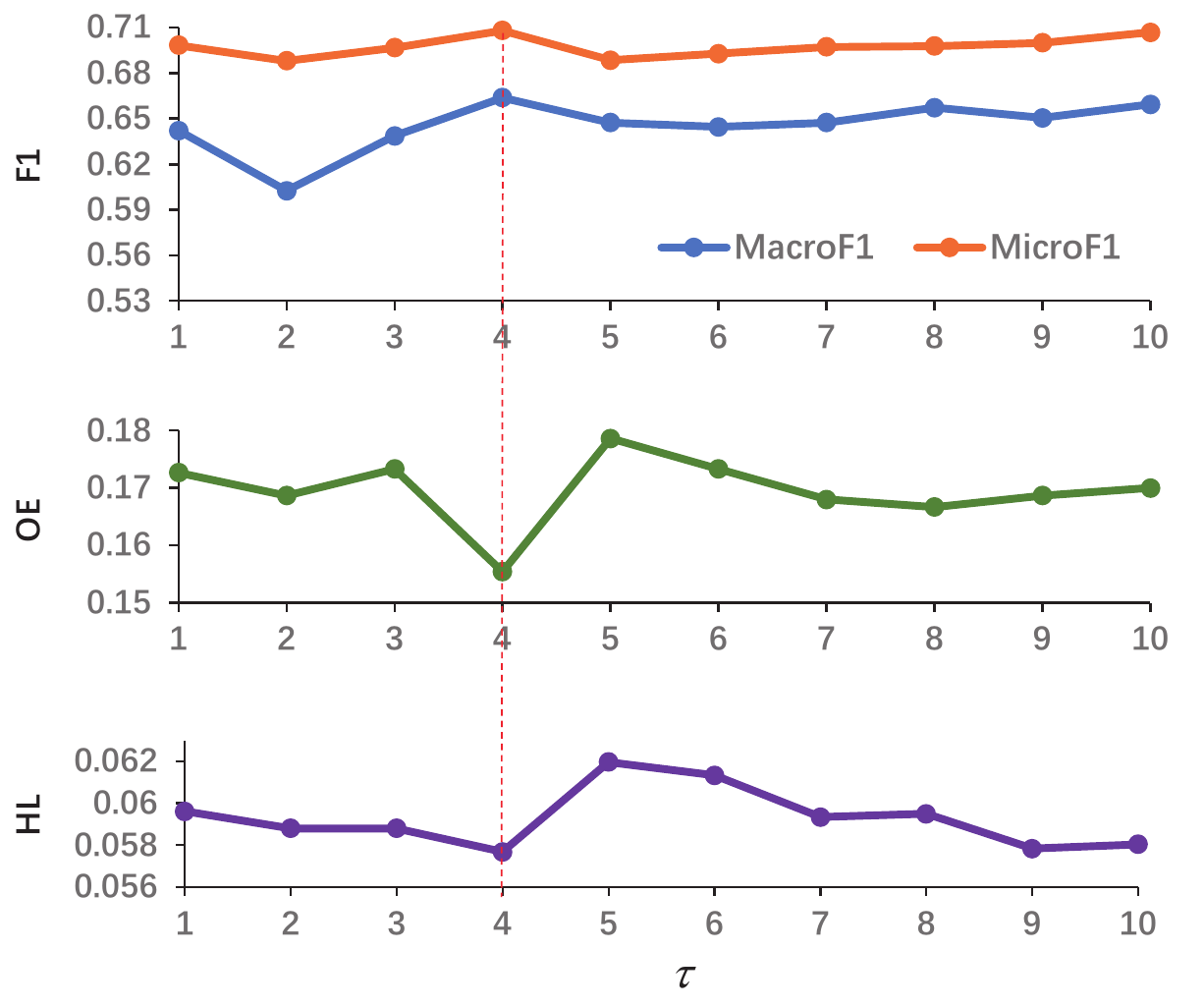} 
	\caption{Parameter sensitivity analysis on the Douban dataset.}
	\label{parameter_sensitivity_analysis}
\end{figure}

The threshold $\tau$ in formula~(\ref{data_correlation_matrix}) may influence the performance. Thus, we conduct a series of sensitivity analysis experiments to study how different choices of parameters influence the performance of the framework and show the results in Figure~\ref{parameter_sensitivity_analysis}. Due to space limitation, we only show our results on Douban dataset.

The threshold $\tau$ decides how many co-occurrence edges in the statistical correlation matrix would be filtered out. It is easy to see that when $\tau$ is too small, the statistical correlation will contain too many rare co-occurrences, which may be noise and cannot reflect the actual co-occurrence patterns of labels. As shown, when filtering out the edges of small probabilities (i.e., noisy edges), the multi-label classification accuracy is boosted. However, when too many useful co-occurrences are filtered out, the performance drops since correlated co-occurrences will be ignored as well. We can observe that \textbf{KRF} achieves the best performance when $\tau = 4$.

\section{Conclusion}
In this paper, we propose a GCN based framework KRF to capture style correlations, which jointly exploits knowledge graph and co-occurrence patterns to derive proper and complete dependencies between different style labels. We first design an effective style correlation matrix by integrating statistical relations and knowledge relations. Then, to capture the deep correlations between labels, we employ GCN to learn label representations using the above style correlation matrix to guide the message propagation among nodes in GCN. Finally, we fuse the label representations and the review representations learned from HAN for music style classification. Extensive experiments conducted on two real-world datasets show that the proposed framework significantly outperforms other state-of-the-art models on multi-label music style classification task and also demonstrates much higher effectiveness to alleviate the underfitting problem.

In the future work, we plan to integrate other types of information such as user preference and artist information into the style correlations. We will explore more efficient methods to learn the label representations for further enhancing the label dependencies learning.

\section{Acknowledgement}
We gratefully thank the anonymous reviewers for their insightful comments. This research is supported in part by the National Key Research and Development Program of China under Grant 2018YFC0806900 and 2017YFB1010000, the Beijing Municipal Science and Technology Project under Grant Z191100007119008. 


\bibliographystyle{ACM-Reference-Format}
\bibliography{main.bib}


\end{document}